%% file: main.tex
\documentclass{article}

\usepackage[ruled]{algorithm2e}
\usepackage{graphicx} 
\usepackage{subcaption}
\usepackage{wrapfig}

\input{math_commands.tex}

\usepackage{dsfont}
\usepackage{hyperref}
\usepackage{url}
\renewcommand{\S}{\mathcal{S}}

\newcommand{\A}{\mathcal{A}}
\newcommand{\EE}{\mathbb{E}}
\newcommand{\sset}{\mathcal{S}}
\newcommand{\aset}{\mathcal{A}}

\newcommand{\trans}{\mathcal{P}}

\usepackage[nonatbib,final]{infer2control_2018}
\usepackage[authoryear]{natbib}
\usepackage[utf8]{inputenc} 
\usepackage[T1]{fontenc}    
\usepackage{hyperref}       
\usepackage{url}            
\usepackage{booktabs}       
\usepackage{amsfonts}       
\usepackage{nicefrac}       
\usepackage{microtype}      

\title{Self-supervised Learning of Image Embedding for Continuous Control}

\author{
Carlos Florensa\thanks{Work done during a summer internship at DeepMind}\\
University of California, Berkeley,\\
florensa@berkeley.edu\\
\And
Jonas Degrave\\
DeepMind\\
London, N1C FCQ, UK \\
\texttt{grave@google.com} \\
\And
Nicolas Heess\\
DeepMind\\
London, N1C FCQ, UK \\
\texttt{heess@google.com} \\
\And
Jost Tobias Springenberg\\
DeepMind\\
London, N1C FCQ, UK \\
\texttt{springenberg@google.com} \\
\And
Martin Riedmiller  \\
DeepMind\\
London, N1C FCQ, UK \\
\texttt{riedmiller@google.com} \\
}

\begin{document}

\maketitle

\begin{abstract}

Operating directly from raw high dimensional sensory inputs like images is still a challenge for robotic control.
Recently, Reinforcement Learning methods have been proposed to solve specific tasks  end-to-end, from pixels to torques.
However, these approaches assume the access to a specified reward which may require specialized instrumentation of the environment. Furthermore, the obtained policy and representations tend to be task specific and may not transfer well.
In this work we investigate completely self-supervised learning of a general image embedding and control primitives, based on finding the shortest time to reach any state. We also introduce a new structure for the state-action value function that builds a connection between model-free and model-based methods, and improves the performance of the learning algorithm. We experimentally demonstrate these findings in three simulated robotic tasks.
\end{abstract}

\section{Introduction}

Reinforcement Learning methods have shown great success in learning complex behaviors, like robotic locomotion \citep{Heess2017emergence,florensa2017snn}, manipulation of objects \citep{florensa2017reverse,Rajeswaran2017dexterous}, or super-human game strategies \citep{mnih2015human,silver2016mastering}. These approaches hold the promise of end-to-end learning of policies that select the optimal action to apply at every state, only leveraging direct interaction with the environment, and a reward signal indicating the desired behavior. Nevertheless, in continuous control many of these approaches still avoid learning from raw sensory input like camera images, or have a separately engineered vision module trained to provide some state information that is known to be relevant for the task \citep{andrychowicz2018dexterity}. This is a considerable drawback to scale up robotic learning and bring them out of instrumented lab environments.

In this work we ask ourselves what a robot could learn solely through self-supervised interaction with the environment, and its high-dimensional sensory input, without any further instrumentation or reward. We show that we can learn a goal-reaching value function \citep{schaul2015uvfa} as well as a corresponding policy that is conditioned on a previously collected observation, and is able to bring the system to a state where the observation closely matches the desired one. This policy is able to connect states even if it has never attempted to reach one from the other.
It could later be used in a hierarchical fashion to perform sequence of operations (like following a desired path, or an assembly instruction manual), or for more meaningful exploration in other downstream tasks \citep{Hausman2018-cp}.

The underlying idea of our algorithm is to learn a state representation where the Euclidean distance between two embedded states corresponds to the minimum number of actions or time-steps needed to connect them. Such an embedding must satisfy a recursive formula that corresponds to the Bellman equation when the reward is the indicator function signaling a perfect match of the observation and the goal. This reward requires no additional instrumentation, but unfortunately the probability of observing it is negligible for any high-dimensional sensory input. To sidestep this issue we can re-label the trajectories as trying to recreate an observation that happened along the trajectory \citep{andrychowicz2017her}, therefore observing this reward as often as needed. This idea was originally studied with low-dimensional state representations, and required a suitably defined distance function to give rewards based on $\epsilon$-Balls around the goals. 
Here, we show that both limitations can be lifted. Furthermore, we introduce a novel architecture for the state-action value function that ties this model-free Reinforcement Learning technique to model-based counterparts.

\section{Background}
In this section we formally introduce the Reinforcement Learning framework, its extension for goal-oriented tasks, and efficient model-free learning algorithms that leverage goal relabeling. We also review the basics of model-based learning as we will draw a connection between the two later on.

\subsection{Reinforcement Learning framework}
We define a discrete-time finite-horizon Markov decision process (MDP) by a tuple $M = (\sset, \aset, \trans, r, \rho_0, T)$, in which $\sset$ is a state set, $\aset$ an action set, $\trans: \sset \times \aset \times \sset \rightarrow \mathbb{R}_{+}$ is a transition probability distribution, $r: \sset \times \aset \rightarrow \R$ is a bounded reward function, $\rho_0: \sset \to \mathbb{R}_+$ is a start state distribution, and $T$ is the horizon. Our aim is to learn a stochastic policy $\pi_{\theta}: \sset \times \aset \to \mathbb{R}_+$ parametrized by $\theta$ that maximizes the expected return, $ \eta_{\rho_0}(\pi_\theta) = \EE_{s_0\sim\rho_0} R(\pi, s_0)$. We denote by $ R(\pi, s_0) := \EE_{\tau|s_0}[ \sum_{t=0}^T r(s_t, a_t) ]$ the expected cumulative reward starting when starting from a $s_0\sim \rho_0$, where $\tau = (s_0, a_0, , \ldots, a_{T-1}, s_T)$ denotes a whole trajectory, with $a_t \sim \pi_\theta(a_t|s_t)$, and $s_{t+1} \sim \trans(s_{t+1} | s_t, a_t)$. 

\subsection{Visual goal tasks}
In this work we assume that we have access to the state only through camera readings, or observations $o_t$. Observations are a function of the current state of the world $s_t$, usually higher dimensional, noisy, and prone to aliasing, like robotic sensory inputs. Technically observations might not be Markovian, but if this is anyhow critical for the task we can always set as state the collection of all previous observations (see Section \ref{sec:limits} for further discussion).
Our objective is to train a policy $\pi(a_t|o_t, o_g)$ that is conditioned on the current observation and a goal observation $o_g$, such that its sampled actions modify the world to obtain an observation that matches $o_g$. 
We employ ideas from Universal Value Function Approximators \citep{schaul2015uvfa} but for tasks in continuous action spaces and where the goals are in the same space as the observations.

\subsection{Off-policy training with goal re-labeling}
\label{sec:her}
When we execute a policy conditioned on a certain goal, it might be that the policy fails to reach that goal, hence does not receive any reward. And if no reward is ever observed, no learning can take place. Fortunately, the goal only affects the reward, but not the dynamics of the environment. Therefore, if the training is done with an off-policy RL algorithm like DDPG \citep{lillicrap2015continuous}, we can re-label \textit{in hindsight} the trajectories to learn about other goals than the one we were originally trying to achieve \citep{andrychowicz2017her}. In particular, if using as goal a point along the trajectory, we are sure to observe a reward.

\subsection{Model-Based Learning}
A forward model is a function $f: \mathcal{O}\times \A\rightarrow\mathcal{O}$ that predicts the next observation from the current observation and action. A standard way to learn such a model is to fit a parameterized function $f_\theta$ by solving
\begin{equation}
\min_{\theta} \sum_{i,t}\|f_{\theta}(o^i_t, a^i_t) - o^i_{t+1}\|, \label{eq:forward_model_fitting}
\end{equation}
where $\big\{(o^i_t, a^i_t, o^i_{t+1})\big\}_{i,t}$ are observed transitions of different trajectories $\tau^i$. Once such a model has been learned, it could be used by search and planning algorithms to select actions that reach a given goal observation $o_g$. One approach would be  Model Predictive Control (MPC) methods \citep{Rao2009control,Nagabandi2017-vs}, where a full sequence of $T$ actions is selected based on the solution to
\begin{equation}
\min_{a_0,\dots, a_T} \|f(f(f(o_0,a_0),a_1)\dots), a_T) - o_g\|, \label{eq:mpc}
\end{equation}
Despite considerable progress in the area \citep{Finn2016unsupervised,Lee2018-ik}, it is still challenging to learn a visual model useful for long-horizon planning, specially due to the intractability of solving \eqref{eq:mpc} exactly, the model-bias and compounding errors in higher dimensions.


\section{Related work}
One of the closes recent work is Temporal Difference Models \citep{Pong2018tdm}, where the authors also try to link model-based and model-free RL through the state-action value function of trying to reach all states. Nevertheless, instead of using directly the minimum number of time-steps to reach a state as we do, they fit the distance at which they will be from the intended goal after $h$ time-steps (with $h$ also given as input to the Q function). Therefore they need to define a distance metric, which is exactly what we try to avoid in our work. Furthermore, this limitation makes them work only from state-space (i.e. positions), where the Euclidean distance is more informative than in image space. Finally, although they sketch a similar connection between MB and MF, theirs only holds for $\gamma=0$, which is not the discount used in practice.

Several work learn a representations of the image observations with some auto-encoding technique \citep{Lange2010-kt}. For example, in \citep{Finn2015autoencoders} an auto-encoder with spatial soft-max is learned on data from trajectories that already succeed at a task 10\% of the time, and then its features are used as the state in a GPS algorithm \citep{Levine2014-qx} to solve that specific task. On the other hand, Visual RL from Imagined Goals \citep{Nair2018rig} has the same objective of trying to reach any reachable observation. They do so training a $\beta$-VAE \citep{Higgins2016-oj} and directly using a distance in that learned space. They obtain interesting results, despite their encoding not reflecting anyhow the dynamics of the environment not what states are actually close to each other in terms of minimum number of actions needed to join them.

Other recent methods have proposed to learn an image embedding spaces where the planning problem could be easier \citep{Kurutach2018-eh,zhang2018solar}. All these methods require more complex methods to learn the embedding than just replacing $o_t$ by $\phi(o_t)$ in \eqref{eq:forward_model_fitting}. Only performing this change would just lead to collapse of the embedding space ($\phi(\cdot)=0$, $f(\cdot, \cdot)=0$ trivially satisfies the equation). 

Successor features \citep{Dayan1993-qs} or representations \citep{Barreto2018-dj,Barreto2016-mj} constitute another related body of work. They also tackle the problem of using the collected transitions for learning policies that can perform well under many rewards. Unfortunately such rewards need to be expressed as a linear combination of some features, and the collection of all indicator rewards cannot be expressed with a finite number of such features. See Appendix \ref{sec:successor} for more details.

\section{Method}

In an MDP setting, two states are considered ``nearby'' if a small number of actions are required to reach one from the other. Unfortunately, a simple distance between states, like the Euclidean, might be uninformative about how nearby two states are.
Most prior work in Reinforcement Learning assumes access to a carefully defined state representation such that a reward based on a distance in that space provides enough shaping to learn the desired policy. The problem is considerably harder when we only have access to high dimensional observations, like images.

Our aim is to learn an embedding space where the distance between embedded observations is representative of (e.g. proportional to) the minimum number of time-steps needed to reach one observation from any other. The main idea is to leverage all collected trajectories to learn such embedding, and, as a by-product, obtain a policy able to reach any past observation upon request. In this section we first recast this as an RL problem, and explain a goal relabeling strategy to solve it. Then we introduce a novel structure for the Q function, that can be seen as bridging the gap between model-based and model-free RL.

\subsection{Minimum time to observation as RL}
Knowing the minimum number of time-steps needed to reach a desired observation as a function of the current state and action is arguably sufficient to perform goal reaching tasks.
A discount-based equivalent of this function can be defined by the recursive equation
\begin{equation}
Q(o_t, a_t, o_g) = \mathds{E}_{o_{t+1}\sim p(\cdot|o_t, a_t)} \Big[\mathds{1}\{o_{t+1} == o_g\} + \mathds{1}\{o_t\neq o_g\}\gamma\max_a Q(o_{t+1},a,o_g) \Big]. \label{eq:q_bellman}
\end{equation}
If the environment is deterministic and the transition $(o_t, a_t, o_{t+1})$ exists, then $Q(o_t, a_t, o_g=o_{t+1})=1$. Equivalently, if at least $h$ steps are needed to reach a certain state $o_{t+h}$ from $o_t$, then we have $Q(o_t, a_t, o_{t+h})=\gamma^h$. Note that the equation above is exactly the Bellman equation defining the optimal $Q$-function \citep{sutton1998reinforcement} for the reward $r(o_t, a_t, o_{t+1}, o_g) = \mathds{1}\{o_{t+1} == o_g\}$. In continuous observation spaces (or discrete but high-dimensional) this reward is not practical 
since even a near-optimal policy might never reach exactly the observation that it is trying to achieve. Therefore to train this value function we use the relabeling strategy outlined in section \ref{sec:her}. In the next section we describe the specific algorithm we use.

\subsection{Goal relabeling and Q fitting}

The ideas outlined in the previous section are not tied to any particular algorithm choice for learning $Q$.
For our experiments we use a goal-conditioned variant of MPO \citep{Abdolmaleki2018mpo}. This algorithm combines a Q fitting done with Retrace \citep{Munos2016retrace} to propagate the discounted returns faster, and a policy improvement step. See details in the Appendix \ref{sec:app_q}.

\begin{algorithm}[H]
\SetAlgoLined
 \SetKwInOut{Input}{Input}
 \SetKwInOut{Output}{Output}
 \Input{Discount $\gamma$, parameterized policy $\pi_{\theta}$, action-value function $Q_{\eta}$, Sequence size $h$, replay buffer $\mathcal{R}$, hindsight $p_g$, Batch Size $B$}
 
 \While{True}{
 	  Sample $(o_t, a_t, r_t, \dots, o_{t+h}, r_{t+h}|o_g) \sim \mathcal{R}$ \tcp*[r]{$o_g$ was the goal observation}
      Sample $ p \sim {\rm Unif}([0,1])$    \tcp*[r]{Use achieved observations as goals}
      \If{$p<p_g$}{
      	Sample $k\sim {\rm Unif}(\{1, 2, \dots, h\})$\;
        Set $o_g=o_{t+k}$ and $r_{t+k+1} = 1$ \tcp*[r]{All other $r_s=0$ already}
      }
      Minimize the loss in \eqref{eq:retrace}\;
      Improve policy by solving \eqref{eq:mpo_improvement}\;
 }
\label{algo:learning_loop}
\caption{\texttt{Learning Loop}}
\end{algorithm}

The learning loop of our full algorithm is described in Algorithm \ref{algo:learning_loop}. Note that there are three sample procedures involved: first obtain a trajectory from the replay buffer, then decide to relabel the trajectory with probability $p_g$ (we use 0.5, but we have seen that, surprisingly, any value between 0.2 and 0.8 yields similar results!). Finally, if we are relabeling the trajectory, we pick a time-step within $h=32$ in the future and use the observation at that time as the goal (also replacing the reward at that time).

The data collection loop executes the most recent policy $\pi_{\theta}$ by conditioning it on any previously observed state as goal. In the next subsection we motivate a new architecture for the Q function, and outline a connection with model learning.

\subsection{Q structure and model-learning}
\begin{figure}[ht]
\captionsetup[subfigure]{justification=centering}
    \centering
      \begin{subfigure}{0.48\linewidth}
        \includegraphics[width=\linewidth, trim={0cm, 0cm, 0cm, 0cm}, clip]{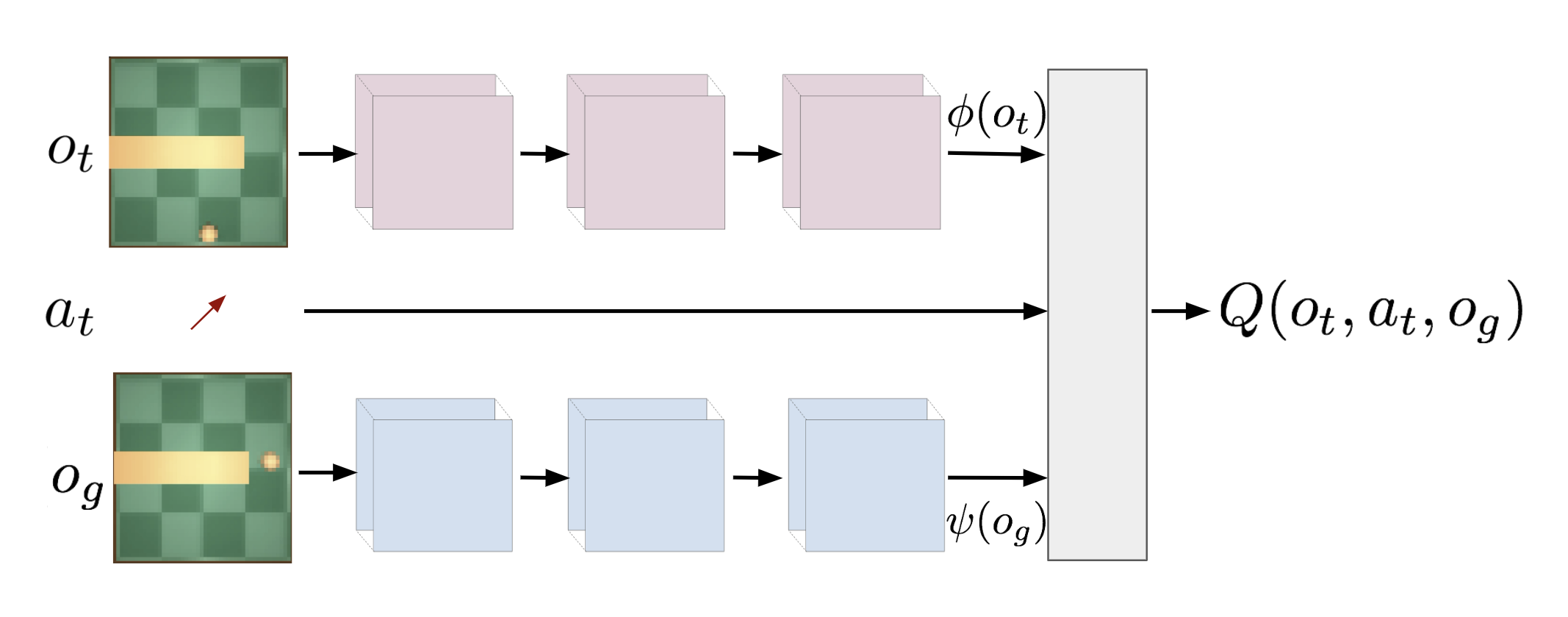}
          \caption{Q unstructured}
          \label{fig:q_unstructured}
      \end{subfigure}
      \unskip\ \vrule\
      \begin{subfigure}{0.48\linewidth}
        \includegraphics[width=\linewidth, trim={0cm, 0cm, 0cm, 0cm}, clip]{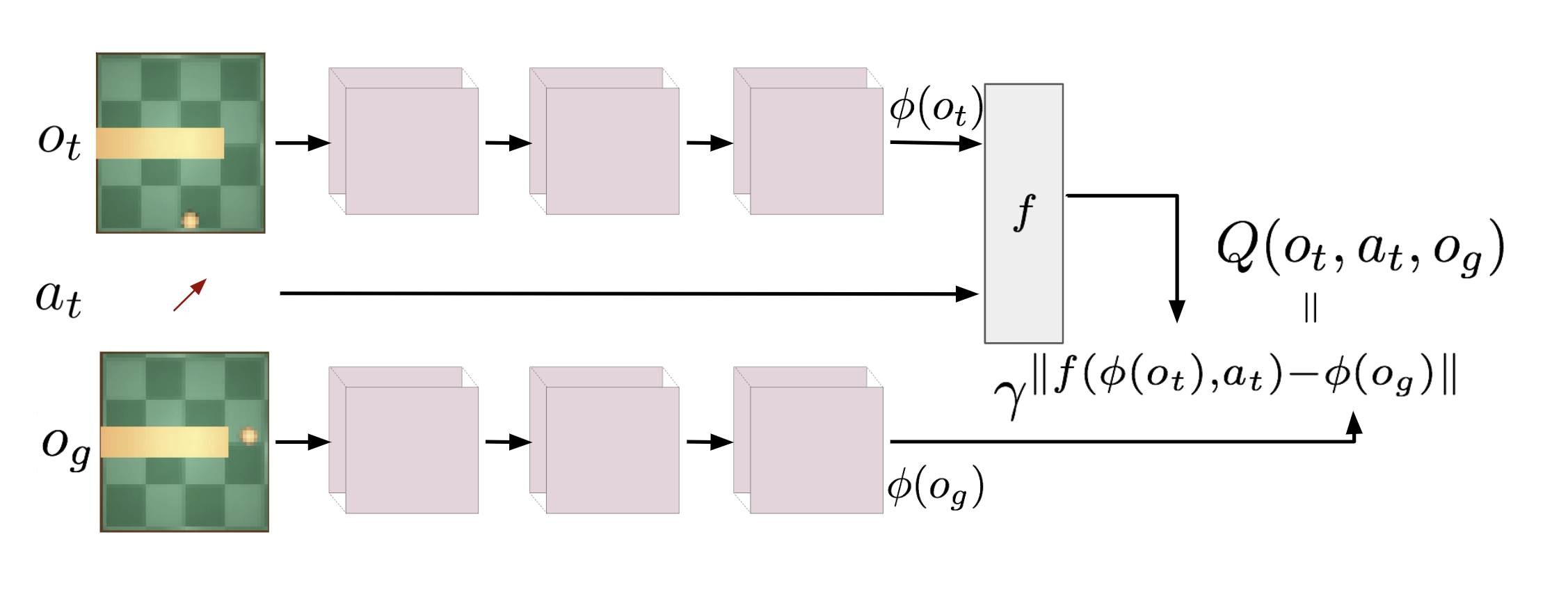}
          \caption{Q structured}
          \label{fig:q_structured}
      \end{subfigure}
\caption{Two Q-function architectures we compare to learn a visual goal-reaching policy.} 
\label{fig:architectures}
\end{figure}

A general architecture for the $Q$ would be to have two independent feature extractors, one for the current observation and another for the goal observation, followed by a Multi-Layer Perceptron (MLP) acting on the concatenated representations. Such an architecture is depicted in Fig.~\ref{fig:q_unstructured}. Is there a more sensible choice of structure for $Q$? First of all, see that this is an universal value function \citep{schaul2015uvfa} that depends on a goal, and in this case the goal also belongs to the observation space. Therefore it is reasonable to apply the same processing to both inputs. Furthermore, the action has no effect on the goal, it only affect the current state. Finally, the Q function should be positive everywhere and evolve exponentially in the minimum number of time-steps needed to reach a state. This suggest the following architecture:
\begin{equation}
Q(o_t, a_t, o_g) = \gamma^{\|f(\phi(o_t), a_t) - \phi(o_g)\|} \label{eq:structured_q}
\end{equation}
where $f$ and $\phi$ are parameterized functions. A scheme of our proposed architecture can be found in Fig.~\ref{fig:q_structured}.

Given the observations from the previous subsections, we realize that using this architecture to optimize \eqref{eq:q_bellman} enforces:
\begin{equation}
\|f(\phi(o_t), a_t) - \phi(o_{t+1})\| = 0 \iff f(\phi(o_t), a_t) = \phi(o_{t+1}). \label{eq:embedded_dynamics}
\end{equation}
Therefore $f$ can be understood as a model in embedding space. Critically, this is not the only equation being fitted, otherwise the embedding would tend to collapse (\eqref{eq:embedded_dynamics} is trivially satisfied by $f(\cdot, \cdot)=0$,  $\phi(\cdot)=0$). Indeed $Q$ is also trained to satisfy \eqref{eq:q_bellman} which enforces $\|f(\phi(o_t), a_t) - \phi(o_{t+h})\| = h$ when $h$ steps are the minimum needed to reach $o_{t+h}$ from $o_t$. Assuming that \eqref{eq:embedded_dynamics} holds true, we are imposing that the embedding satisfies $\|\phi(o_{t+1}) - \phi(o_{t+h})\| = h$. Therefore this can be understood as a model enhanced with stronger planning capabilities, also giving an embedding where distances are proportional to shortest paths between points.

As an additional observation, the sample efficiency of model-based RL is often attributed to the ability to make use of all observed data, as any valid transition is informative of the dynamics. 
Relabeling procedures as the one used here achieves a similar effect, and along with the specific structure introduced above we blur the lines between both types of methods.

\section{Experiments}

\begin{figure}[ht]
\captionsetup[subfigure]{justification=centering}
    \centering
      \begin{subfigure}{0.25\linewidth}
        \includegraphics[width=\linewidth, trim={0cm, 0cm, 0cm, 0cm}, clip]{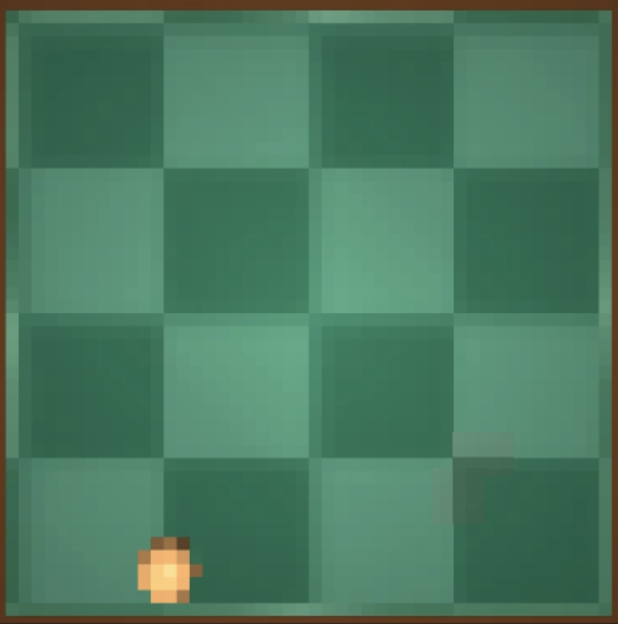}
          \caption{Point mass}
          \label{fig:pt}
      \end{subfigure}
      \begin{subfigure}{0.25\linewidth}
        \includegraphics[width=\linewidth, trim={0cm, 0cm, 0cm, 0cm}, clip]{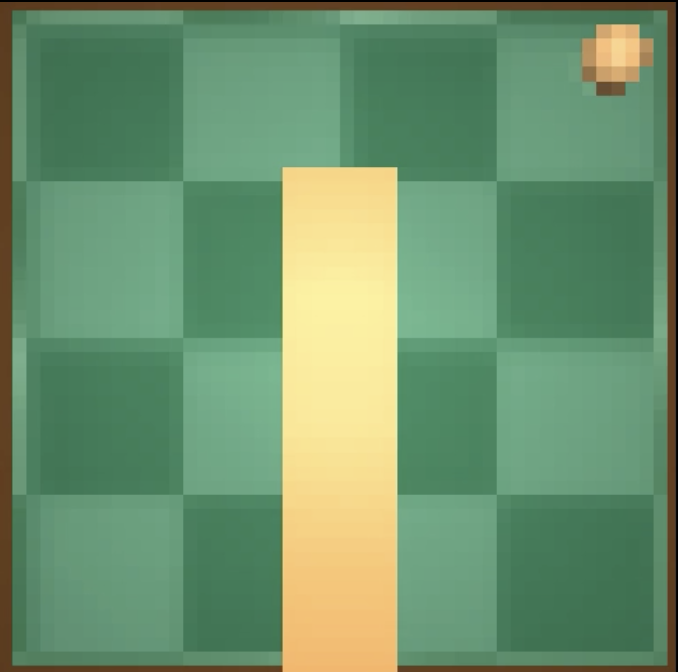}
          \caption{Wall point mass}
          \label{fig:wall}
      \end{subfigure}
      \begin{subfigure}{0.25\linewidth}
        \includegraphics[width=\linewidth, trim={0cm, 0cm, 0cm, 0cm}, clip]{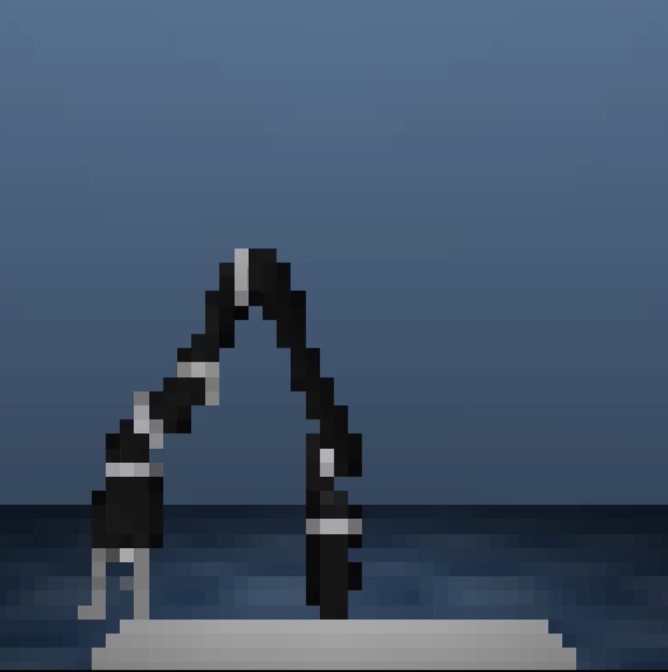}
          \caption{Jaco arm reacher}
          \label{fig:arm}
      \end{subfigure}
\caption{Task observation, at the resolution given to the agent. No other proprioceptive or geometric information is used. The goal is also specified as an observation like the above.} 
\label{fig:tasks}
\end{figure}

We investigate the following questions: 1) Can we learn a goal-reaching policy solely from visual input and no rewards? 2) Does adding structure to the Q function improve the performance of the algorithm? 3) Does the learned embedding carry dynamics information like time-steps between points along a trajectory? We analyze this questions in three environments implemented with the physics simulator MuJoCo \citep{Todorov2012mujoco}: visual point-mass, wall point-mass, and Jaco arm reacher. The first two environments have a two-dimensional action-space corresponding to a force applied to a spherical object. For both the observation is a fully top-down view of the scene, as seen in Fig.~\ref{fig:pt}-\ref{fig:wall}. The only difference is a wall blocking 3/4 of the middle division in the wall task, creating a sort of U-maze. The Jaco arm has a seven-dimensional action-space to control the velocity of each joint actuator. The observation is a frontal view. All trajectories last 10 seconds and the agents operate at 10Hz for the point-mass and at 20Hz for the Jaco. Our algorithm works solely from visual inputs with the resolution observed in Fig.~\ref{fig:tasks}: 64x64 pixels for the point-mass environments and 96x96 pixels for the Jaco arm.

\subsection{Self-supervised learning image-conditioned policies}

In this subsection we show that we can learn goal-conditioned policies that control the agent to reach a state which observation matches any previously seen goal observation. No reward, nor state information, is ever used in the learning process. Nevertheless, we use the L1 distance in position-space as learning progress metric given that distances in pixel-space are more noisy and less interpretable. For the point-mass tasks, the position is the $(x,y)$ coordinates of the Center of Mass. For the Jaco arm, it is the seven-dimensional joint angles.

\begin{figure}[ht]
\captionsetup[subfigure]{justification=centering}
    \centering
      \begin{subfigure}{0.32\linewidth}
        \includegraphics[width=\linewidth, trim={0cm, 0cm, 0cm, 0cm}, clip]{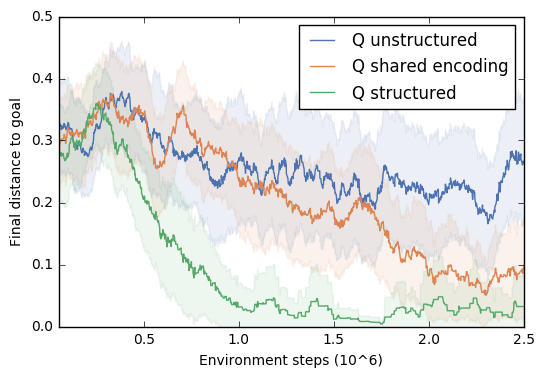}
          \caption{Point mass}
          \label{fig:learning_pt}
      \end{subfigure}
      \begin{subfigure}{0.32\linewidth}
        \includegraphics[width=\linewidth, trim={0cm, 0cm, 0cm, 0cm}, clip]{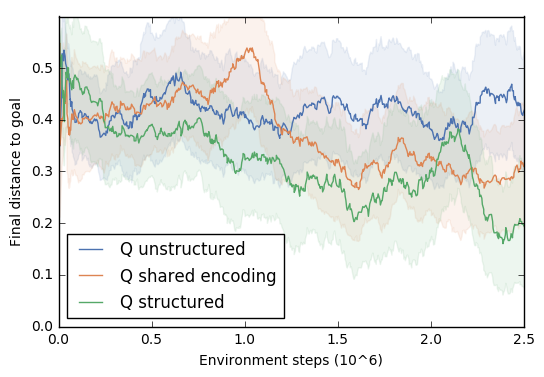}
          \caption{Wall point mass}
          \label{fig:learning_wall}
      \end{subfigure}
      \begin{subfigure}{0.32\linewidth}
        \includegraphics[width=\linewidth, trim={0cm, 0cm, 0cm, 0cm}, clip]{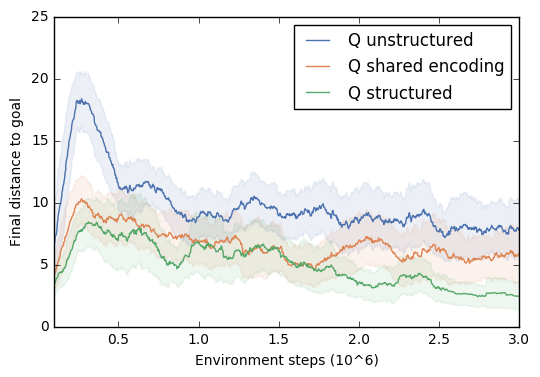}
          \caption{Jaco arm reacher}
          \label{fig:learning_arm_task}
      \end{subfigure}
\caption{Learning curves for the three environments plotting final L1 goal distance in position-space against collected environment steps.} 
\label{fig:learning}
\end{figure}

In Fig.~\ref{fig:learning} we compare three versions of the algorithm, differing only in the structure of the Q-function. \textit{Q unstructured} does not impose any structure on the $Q$ function, \textit{Q shared encoding} uses the same vision stack to process the current observation and the goal observation, and \textit{Q structured} additionally imposes the structure given in \eqref{eq:structured_q}. 
To answer our first question, we observe that the algorithm is able to reduce the final distance to the given goal using any of the three models. The performance reported in these plots is computed based on collecting trajectories conditioned on previously seen observations as goals. Therefore, as the replay buffer grows, the evaluation criteria gets harder at the start of learning, specially for the higher dimensional environments like the Jaco.
To answer our second question,  we see that the structure that we introduced in the previous section substantially increases convergence speed and final performance attained.

\subsection{Embedding analysis}
\begin{figure}
    \centering
    \includegraphics[width=\linewidth]{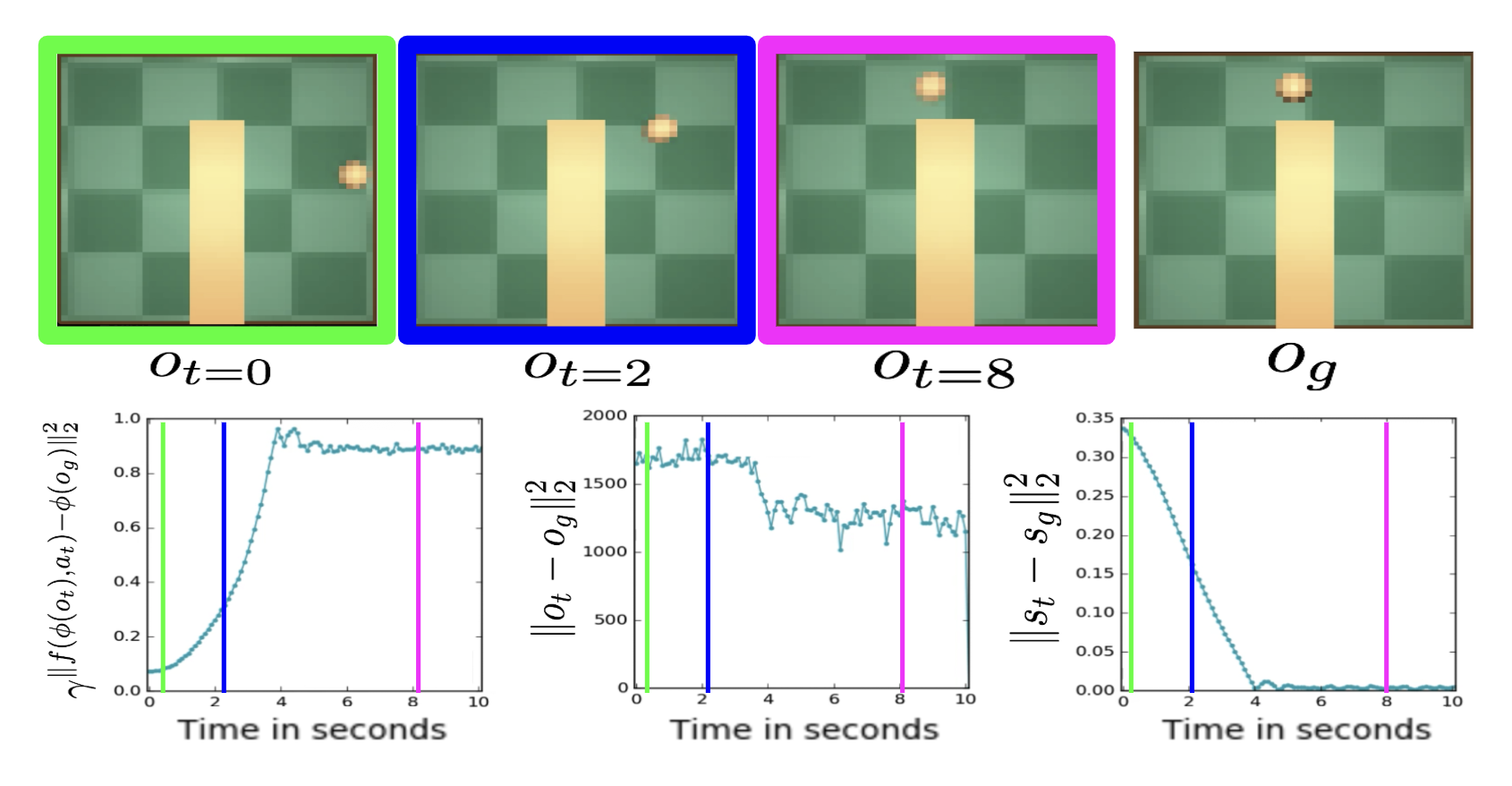}
    \caption{Analysis of some distances along a trajectory}
    \label{fig:wall_traj}
\end{figure}

In this section we study the evolution of three types of distances to the goal along a successful trajectory for the Wall point-mass task. Similar plots can be found for the other environments. In the top row of Fig.~\ref{fig:wall_traj} we observe three frames obtained at $t=$ 0, 2, and 8 seconds, as well as the goal image $o_g$. In the bottom row we monitor, from left to right, our structured $Q$ function, the distance in pixel space and the distance in position space. From the left figure we see that the agent reaches the exact position that generated the goal observation at $t=4$, and stays there with small oscillations. We see in the pixel distance plot that this is an uninformative distance before having reached the vicinity of the goal (all observations before $t=4$ seconds are at the same noisy distance), and even after reaching the goal it is never reduced to 0 because the observations never match exactly. Therefore it is hard to interpret or use this distance as a reward to learn a goal-reaching policy. Finally, in the left figure we see that the distance in embedding space trained through our structured $Q$ follows $\gamma^h$, where $h$ is the remaining number of time-steps to the goal. The fact that it does not reach exactly 1 is because it never exactly reaches the same observation, but it understands that only a few time-steps would be needed (in theory, as in practice it will never match the exact same observation).

\section{Failure modes and future work}
\label{sec:limits}
Both in the Wall point-mass, and the Jaco arm environments we do not obtain a complete convergence with our algorithm. Here we describe some existing issues, suggest an hypothesis about their source, and some experiments to check their validity.

First, in all the environments we observe some oscillation around the goal position. For the point-mass environments this is not critical, but for the Jaco arm we have observed that this is the cause of most of the final distance to goal. 
We know that in many robotic environments, a knowledge of the velocities is critical to act optimally to reach specified states (like estimating if the agent is going in the right direction). Nevertheless this information cannot be conveyed in a single image. Furthermore, working from images directly may introduce issues with observation aliasing when several states produce similar observations. Both problems could be mitigated by adding as input to the $Q$ and $\pi$ the observations from some previous time-steps. For example, the model could be $f(\phi(o_t), \phi(o_{t-1}), a_t)$, or be an RNN taking in all previous observations. Another solution to the oscillation problem would be a change in the action space: if the system allows the use of delta-position commands this would greatly alleviate the issue. 

Second, in the wall point-mass we observe some difficulties in reaching goals that are very far into the other leg of the U-shape. We think this might be an exploration problem. Indeed our method completely overlooks this issue (ie, is orthogonal to it), solely relying on the random initialization, and on the maximum entropy policy given by MPO. This might not give enough structure to the space, specifically to link far away states. To alleviate this issue we could either use some intrinsic motivation reward to expand the set of observed goals, or add to the replay buffer some demonstrations performing the hardest connections.

Finally, we would like to point out at some limitations of the exact formulation we propose here, and possible fixes. An underlying assumption of our work is that the environment is reversible, otherwise there is no embedding space where a distance (which by definition is symmetric) can be equal to the minimum time-steps between states. This is true for a wide range of practical tasks like all quasi-static manipulation, but might not hold when acting on deformable objects or highly dynamic tasks like throwing objects. In such cases, we should replace the distance in the Q function model by a non-symmetric comparison between states.

\section{Conclusions}
We have shown it is possible to learn goal-reaching policies in a completely self-supervised setup, and only from high dimensional sensory inputs like images. 
Our approach does not use any auxiliary learning signal. Instead, it relies solely on computing the minimum number of time-steps needed to connect different states. This can be written as a Bellman equation that we efficiently solve with a modified off-policy algorithm paired with goal relabeling.
We also introduce a novel structure of the $Q$ function that connects model-free and model-based RL methods, as well as improving the learning speed and final performance.


\bibliography{refs}

\appendix
\section{Q fitting}
\label{sec:app_q}
 Let's denote by $\eta$ all the parameters of the $Q$ function, $\eta'$ their target values that are updated with the value of $\eta$ every few learning iterations, $b(a|o)$ an arbitrary behavior policy, $\mu_b(o)$ its induced state visitation, and $\pi_{\theta}$ our current policy. Then the value estimation step amounts to solving:
\begin{equation}
\min_{\eta} L(\eta) = \min_{\eta}\E_{\mu_b(o), b(a|o), o_g}\big[\big(Q_{\eta}(o_t,a_t, o_g) - Q^{ret}_{t, o_g} \big)^2 \big], {\qquad {\rm where }}\label{eq:retrace}
\end{equation}
\begin{equation*}
Q^{ret}_{t,o_g} = Q_{\eta'}(o_t, a_t, o_g) + \sum_{s=t}^{t+h}\gamma^{s-t}\Big(\prod_{i=t+1}^s c_i\Big)\big[ r_s + \gamma \E_{a\sim\pi_{\theta}(\cdot|o_{s+1},o_g)} Q_{\eta'}(o_{s+1},\cdot, o_g) - Q_{\eta'}(o_s, a_s, o_g)\big]
\end{equation*}
\begin{equation*}
c_i = \min\Big(1, \frac{\pi_{\theta}(a_s|o_s, o_g)}{b(a_s|o_s, o_g)}\Big)
\end{equation*}
Once an approximation of the $Q$ value of the current policy $\pi_{\theta}$ is known, we can express the policy that maximizes $\E_{\mu(o)}\E_q(a|o)[Q_{\theta_i}(o,a)]$, under the trust region $\E_{\mu(o)}KL\big(q(a|o), \pi(a|o, \theta_i)\big) \leq \epsilon$ as $$q(a|o) \propto \pi_{\theta_i}(a|o)\exp(Q_{\theta_i}(o,a) / \tau),$$
where the value of $\tau$ is the solution of a convex dual problem \citep{Peters2010-ra}. This is a non-parametric form, so to recover a policy $\pi_{\theta_{i+1}}$ from where we can sample, we can solve another KL-constrained maximum likelihood problem, where $\lambda$ is the corresponding dual variable \citep{Abdolmaleki2018mpo}:
\begin{equation}
\max_{\theta_{i+1}}\E_{\mu_q(o)}\E_{q(a|o)}\big[\log\pi_{\theta_{i+1}}(a|o, o_g)\big] - \lambda KL\big(\pi_{\theta_i}(a|o, o_g), \pi_{\theta_{i+1}}(a|o, o_g) \big) \label{eq:mpo_improvement}
\end{equation}

\section{Hyperparameteres used}
\label{sec:hyper}
\subsection{Architecture choices}
The vision stack for the critic consists of five convolutions of strides $(1, 2, 2, 2, 2)$ and output channels $(8, 16, 32, 16, 8)$. All kernel shapes are $(3,3)$.
The number of encoded features (ie, the output dimension of the convolution stack) is 128.
The current representations are then concatenated with the action, and passed through two fully connected layers of sizes $(200, 400)$ with ReLU non-linearities. The final output is a scalar with a tanh nonlinearity.
The policy $\pi_{\theta}$ has the same architecture as the critic, with the exception of the output being of the dimension of twice the action space to parameterize a Gaussian distribution (with diagonal covariance matrix). 

\subsection{Algorithm hyperparameters}
The batch size to form the Retrace loss and the MPO objective consists of 128 sequences of 32 steps from the replay buffer. The MPO objective has an initial temperature of $\tau=0.1$, and a KL constraint of $\epsilon=0.2$. The optimization algorithm to minimize these losses is Adam, with a learning rate of $5^{-4}$. The critic target is updated every 8 learning iterations. The capacity of the replay buffer is set to $10^{5}$ trajectories.

\section{Successor features connection}
\label{sec:successor}
Successor Representations (SR) have been used to train state-reaching tasks in simple continuous cases. Nevertheless there is a major limitation to the previously proposed approaches: all states that we will ever be interested in learning a reaching policy need to be pre-specified before starting any learning! In the case of \citep{Barreto2016-mj}, those are only 12. The reason behind this limitation is that SR require the reward function to be defined as a linear function of some state features $r_w(s) = \phi(s)^Tw$. Therefore, if we want to express rewards related to reaching a particular state $s_g$, like $r(s) = \mathds{1}[s==s_g]$, we need to have a component $i$ in the state representation vector $\phi(s)$ that gives exactly this value, such that the reward can be expressed with $w=e_i$. This is because rewards like the ones specified above cannot be expressed linearly as a function of $s_g$. Unfortunately this trick can only be done a finite number of times, as many as we are willing to increase the dimensionality of $\phi(s)$.

In fact, taking this process to the extreme, the feature "vector" becomes a function $\phi(s,g)=\mathds{1}[s==g]$, and then the reward needs to be expressed as $r_w(s)=\int_{g\in \S}\phi(s,g)w(g)dg = w(s)$. In other words, in this case the ``vector'' $w$ is simply any function of the state, meaning we can represent any reward! This seems to indicate that if we compute the SR for a policy $\pi$ (now also dependent on the goal $g$):
$$\Psi^{\pi}(s,a,g) = \mathds{1}[s_{t+1}==g] + \gamma\E_{a\sim\pi(\cdot|s_{t+1})}\Big[\Psi^{\pi}(s_{t+1}, a, g)\Big],$$
we could directly find the action-value function for any reward: $Q^{\pi}_{r}(s,a) = \int_{g\in\S}\Psi^{\pi}(s,g)r(g)dg$. Of course this is not very practical as computing this integral is probably as hard as computing the Q from scratch.


\end{document}

%% file: math_commands.tex
\usepackage{amsmath,amsfonts,bm}

\def\eqref#1{equation~\ref{#1}}

\def\1{\bm{1}}

\DeclareMathAlphabet{\mathsfit}{\encodingdefault}{\sfdefault}{m}{sl}
\SetMathAlphabet{\mathsfit}{bold}{\encodingdefault}{\sfdefault}{bx}{n}

\newcommand{\E}{\mathbb{E}}

\newcommand{\R}{\mathbb{R}}



%% file: main.bbl
\begin{thebibliography}{31}
\providecommand{\natexlab}[1]{#1}
\providecommand{\url}[1]{\texttt{#1}}
\expandafter\ifx\csname urlstyle\endcsname\relax
  \providecommand{\doi}[1]{doi: #1}\else
  \providecommand{\doi}{doi: \begingroup \urlstyle{rm}\Url}\fi

\bibitem[Abdolmaleki et~al.(2018)Abdolmaleki, Springenberg, Tassa, Munos,
  Heess, and Riedmiller]{Abdolmaleki2018mpo}
Abbas Abdolmaleki, Jost~Tobias Springenberg, Yuval Tassa, Remi Munos, Nicolas
  Heess, and Martin Riedmiller.
\newblock Maximum a posteriori policy optimisation.
\newblock \emph{International Conference on Learning Representations}, 2018.

\bibitem[Andrychowicz et~al.(2017)Andrychowicz, Wolski, Ray, Schneider, Fong,
  Welinder, McGrew, Tobin, Abbeel, and Zaremba]{andrychowicz2017her}
Marcin Andrychowicz, Filip Wolski, Alex Ray, Jonas Schneider, Rachel Fong,
  Peter Welinder, Bob McGrew, Josh Tobin, Pieter Abbeel, and Wojciech Zaremba.
\newblock Hindsight experience replay.
\newblock In \emph{Advances in Neural Information Processing Systems}, 2017.

\bibitem[Andrychowicz et~al.(2018)Andrychowicz, Baker, Chociej, Jozefowicz,
  McGrew, Pachocki, Petron, Plappert, Powell, Ray, Schneider, Sidor, Tobin,
  Welinder, Weng, and Zaremba]{andrychowicz2018dexterity}
Marcin Andrychowicz, Bowen Baker, Maciek Chociej, Rafal Jozefowicz, Bob McGrew,
  Jakub Pachocki, Arthur Petron, Matthias Plappert, Glenn Powell, Alex Ray,
  Jonas Schneider, Szymon Sidor, Josh Tobin, Peter Welinder, Lilian Weng, and
  Wojciech Zaremba.
\newblock Learning dexterous {In-Hand} manipulation.
\newblock In \emph{https://arxiv.org/abs/1808.00177}, 2018.

\bibitem[Barreto et~al.(2016)Barreto, Dabney, Munos, Hunt, Schaul, van Hasselt,
  and Silver]{Barreto2016-mj}
Andr{\'e} Barreto, Will Dabney, R{\'e}mi Munos, Jonathan~J Hunt, Tom Schaul,
  Hado van Hasselt, and David Silver.
\newblock Successor features for transfer in reinforcement learning.
\newblock \emph{http://arxiv.org/abs/1606.05312}, 2016.

\bibitem[Barreto et~al.(2018)Barreto, Borsa, Quan, Schaul, Silver, Hessel,
  Mankowitz, Zidek, and Munos]{Barreto2018-dj}
Andre Barreto, Diana Borsa, John Quan, Tom Schaul, David Silver, Matteo Hessel,
  Daniel Mankowitz, Augustin Zidek, and Remi Munos.
\newblock Transfer in deep reinforcement learning using successor features and
  generalised policy improvement.
\newblock In Jennifer Dy and Andreas Krause, editors, \emph{Proceedings of the
  35th International Conference on Machine Learning}, volume~80 of
  \emph{Proceedings of Machine Learning Research}, pages 501--510,
  Stockholmsm{\"a}ssan, Stockholm Sweden, 2018. PMLR.

\bibitem[Dayan(1993)]{Dayan1993-qs}
Peter Dayan.
\newblock Improving generalization for temporal difference learning: The
  successor representation.
\newblock \emph{Neural Comput.}, 5\penalty0 (4):\penalty0 613--624, 1993.

\bibitem[Finn et~al.(2015)Finn, Tan, Duan, Darrell, Levine, and
  Abbeel]{Finn2015autoencoders}
Chelsea Finn, Xin~Yu Tan, Yan Duan, Trevor Darrell, Sergey Levine, and Pieter
  Abbeel.
\newblock Deep spatial autoencoders for visuomotor learning.
\newblock In \emph{International Conference on Robotics and Automation}, 2015.

\bibitem[Finn et~al.(2016)Finn, Goodfellow, and Levine]{Finn2016unsupervised}
Chelsea Finn, Ian Goodfellow, and Sergey Levine.
\newblock Unsupervised learning for physical interaction through video
  prediction.
\newblock In \emph{Advances in Neural Information Processing Systems}, 2016.

\bibitem[Florensa et~al.(2017{\natexlab{a}})Florensa, Duan, and
  Abbeel]{florensa2017snn}
Carlos Florensa, Yan Duan, and Pieter Abbeel.
\newblock Stochastic neural networks for hierarchical reinforcement learning.
\newblock \emph{International Conference in Learning Representations},
  2017{\natexlab{a}}.

\bibitem[Florensa et~al.(2017{\natexlab{b}})Florensa, Held, Wulfmeier, Zhang,
  and Abbeel]{florensa2017reverse}
Carlos Florensa, David Held, Markus Wulfmeier, Michael Zhang, and Pieter
  Abbeel.
\newblock Reverse curriculum generation for reinforcement learning.
\newblock \emph{Conference on Robot Learning}, 2017{\natexlab{b}}.

\bibitem[Hausman et~al.(2018)Hausman, Springenberg, Wang, Heess, and
  Riedmiller]{Hausman2018-cp}
Karol Hausman, Jost~Tobias Springenberg, Ziyu Wang, Nicolas Heess, and Martin
  Riedmiller.
\newblock Learning an embedding space for transferable robot skills.
\newblock \emph{International Conference on Learning Representations}, 2018.

\bibitem[Heess et~al.(2017)Heess, Dhruva, Sriram, Lemmon, Merel, Wayne, Tassa,
  Erez, Wang, Ali~Eslami, Riedmiller, and Silver]{Heess2017emergence}
Nicolas Heess, T~B Dhruva, Srinivasan Sriram, Jay Lemmon, Josh Merel, Greg
  Wayne, Yuval Tassa, Tom Erez, Ziyu Wang, S~M Ali~Eslami, Martin Riedmiller,
  and David Silver.
\newblock Emergence of locomotion behaviours in rich environments.
\newblock \emph{http://arxiv.org/abs/1707.02286}, 2017.

\bibitem[Higgins et~al.(2016)Higgins, Matthey, Pal, Burgess, Glorot, Botvinick,
  Mohamed, and Lerchner]{Higgins2016-oj}
Irina Higgins, Loic Matthey, Arka Pal, Christopher Burgess, Xavier Glorot,
  Matthew Botvinick, Shakir Mohamed, and Alexander Lerchner.
\newblock {beta-VAE}: Learning basic visual concepts with a constrained
  variational framework.
\newblock In \emph{International Conference in Learning Representations}, 2016.

\bibitem[Kurutach et~al.(2018)Kurutach, Tamar, Yang, Russell, and
  Abbeel]{Kurutach2018-eh}
Thanard Kurutach, Aviv Tamar, Ge~Yang, Stuart Russell, and Pieter Abbeel.
\newblock Learning plannable representations with causal {InfoGAN}.
\newblock \emph{http://arxiv.org/abs/1807.09341}, 2018.

\bibitem[Lange and Riedmiller(2010)]{Lange2010-kt}
Sascha Lange and Martin~A Riedmiller.
\newblock Deep learning of visual control policies.
\newblock In \emph{{ESANN}}, 2010.

\bibitem[Lee et~al.(2018)Lee, Zhang, Ebert, Abbeel, Finn, and
  Levine]{Lee2018-ik}
Alex~X Lee, Richard Zhang, Frederik Ebert, Pieter Abbeel, Chelsea Finn, and
  Sergey Levine.
\newblock Stochastic adversarial video prediction.
\newblock \emph{http://arxiv.org/abs/1804.01523}, 2018.

\bibitem[Levine and Abbeel(2014)]{Levine2014-qx}
Sergey Levine and Pieter Abbeel.
\newblock Learning neural network policies with guided policy search under
  unknown dynamics.
\newblock In \emph{Advances in Neural Information Processing Systems 27}, pages
  1071--1079. Curran Associates, Inc., 2014.

\bibitem[Lillicrap et~al.(2016)Lillicrap, Hunt, Pritzel, Heess, Erez, Tassa,
  Silver, and Wierstra]{lillicrap2015continuous}
Timothy~P Lillicrap, Jonathan~J Hunt, Alexander Pritzel, Nicolas Heess, Tom
  Erez, Yuval Tassa, David Silver, and Daan Wierstra.
\newblock Continuous control with deep reinforcement learning.
\newblock In \emph{International Conference on Learning Representations}, 2016.

\bibitem[Mnih et~al.(2015)Mnih, Kavukcuoglu, Silver, Rusu, Veness, Bellemare,
  Graves, Riedmiller, Fidjeland, Ostrovski, Petersen, Beattie, Sadik,
  Antonoglou, King, Kumaran, Wierstra, Legg, and Hassabis]{mnih2015human}
Volodymyr Mnih, Koray Kavukcuoglu, David Silver, Andrei~A Rusu, Joel Veness,
  Marc~G Bellemare, Alex Graves, Martin Riedmiller, Andreas~K Fidjeland, Georg
  Ostrovski, Stig Petersen, Charles Beattie, Amir Sadik, Ioannis Antonoglou,
  Helen King, Dharshan Kumaran, Daan Wierstra, Shane Legg, and Demis Hassabis.
\newblock Human-level control through deep reinforcement learning.
\newblock \emph{Nature}, 518\penalty0 (7540):\penalty0 529--533, 2015.

\bibitem[Munos et~al.(2016)Munos, Stepleton, Harutyunyan, and
  Bellemare]{Munos2016retrace}
R{\'e}mi Munos, Tom Stepleton, Anna Harutyunyan, and Marc~G Bellemare.
\newblock Safe and efficient {Off-Policy} reinforcement learning.
\newblock In \emph{Advances in Neural Information Processing Systems}, 2016.

\bibitem[Nagabandi et~al.(2017)Nagabandi, Kahn, Fearing, and
  Levine]{Nagabandi2017-vs}
Anusha Nagabandi, Gregory Kahn, Ronald~S Fearing, and Sergey Levine.
\newblock Neural network dynamics for {Model-Based} deep reinforcement learning
  with {Model-Free} {Fine-Tuning}.
\newblock \emph{http://arxiv.org/abs/1708.02596}, 2017.

\bibitem[Nair et~al.(2018)Nair, Pong, Dalal, Bahl, Lin, and
  Levine]{Nair2018rig}
Ashvin Nair, Vitchyr Pong, Murtaza Dalal, Shikhar Bahl, Steven Lin, and Sergey
  Levine.
\newblock Visual reinforcement learning with imagined goals.
\newblock \emph{http://arxiv.org/abs/1807.04742}, 2018.

\bibitem[Peters et~al.(2010)Peters, M{\"u}lling, and Altun]{Peters2010-ra}
J~Peters, K~M{\"u}lling, and Y~Altun.
\newblock Relative entropy policy search.
\newblock \emph{AAAI}, 2010.

\bibitem[Pong et~al.(2018)Pong, Gu, Dalal, and Levine]{Pong2018tdm}
Vitchyr Pong, Shixiang Gu, Murtaza Dalal, and Sergey Levine.
\newblock Temporal difference models: {Model-Free} deep {RL} for {Model-Based}
  control.
\newblock In \emph{International Conference on Learning Representations}, 2018.

\bibitem[Rajeswaran et~al.(2018)Rajeswaran, Kumar, Gupta, Schulman, Todorov,
  and Levine]{Rajeswaran2017dexterous}
Aravind Rajeswaran, Vikash Kumar, Abhishek Gupta, John Schulman, Emanuel
  Todorov, and Sergey Levine.
\newblock Learning complex dexterous manipulation with deep reinforcement
  learning and demonstrations.
\newblock In \emph{Robotics: Science and Systems}, 2018.

\bibitem[Rao(2009)]{Rao2009control}
Anil~V Rao.
\newblock A survey of numerical methods for optimal control.
\newblock \emph{Advances in the Astronautical Sciences}, AAS 09-334, 2009.

\bibitem[Schaul et~al.(2015)Schaul, Horgan, Gregor, and Silver]{schaul2015uvfa}
Tom Schaul, Dan Horgan, Karol Gregor, and David Silver.
\newblock Universal value function approximators.
\newblock In \emph{International Conference on Machine Learning}, 2015.

\bibitem[Silver et~al.(2016)Silver, Huang, Maddison, Guez, Sifre, van~den
  Driessche, Schrittwieser, Antonoglou, Panneershelvam, Lanctot, Dieleman,
  Grewe, Nham, Kalchbrenner, Sutskever, Lillicrap, Leach, Kavukcuoglu, Graepel,
  and Hassabis]{silver2016mastering}
David Silver, Aja Huang, Chris~J Maddison, Arthur Guez, Laurent Sifre, George
  van~den Driessche, Julian Schrittwieser, Ioannis Antonoglou, Veda
  Panneershelvam, Marc Lanctot, Sander Dieleman, Dominik Grewe, John Nham, Nal
  Kalchbrenner, Ilya Sutskever, Timothy Lillicrap, Madeleine Leach, Koray
  Kavukcuoglu, Thore Graepel, and Demis Hassabis.
\newblock Mastering the game of go with deep neural networks and tree search.
\newblock \emph{Nature}, 529\penalty0 (7587):\penalty0 484--489, 2016.

\bibitem[Sutton and Barto(1998)]{sutton1998reinforcement}
Richard~S Sutton and Andrew~G Barto.
\newblock \emph{Reinforcement learning: An introduction}.
\newblock MIT press, 1998.

\bibitem[Todorov et~al.(2012)Todorov, Erez, and Tassa]{Todorov2012mujoco}
E~Todorov, T~Erez, and Y~Tassa.
\newblock {MuJoCo}: A physics engine for model-based control.
\newblock In \emph{2012 {IEEE/RSJ} International Conference on Intelligent
  Robots and Systems}, pages 5026--5033, 2012.

\bibitem[Zhang et~al.(2018)Zhang, Vikram, Smith, Abbeel, Johnson, and
  Levine]{zhang2018solar}
Marvin Zhang, Sharad Vikram, Laura Smith, Pieter Abbeel, Matthew~J Johnson, and
  Sergey Levine.
\newblock Solar: Deep structured latent representations for model-based
  reinforcement learning.
\newblock \emph{http://arxiv.org/abs/1808.09105}, 2018.

\end{thebibliography}
